# Fault-Trajectory Approach for Fault Diagnosis on Analog Circuits

Carlos Eduardo Savioli,
Claudio C. Czendrodi
*Brazilian Navy Electronics Center*
264@cetm.mar.mil.br

Jose Vicente Calvano
*Brazilian Navy Research Institute*
Calvano@olimpo.com.br

Antonio Carneiro de Mesquita Filho
*Federal University of Rio de Janeiro*
Mesquita@coe.ufrj.br

## Abstract

*This issue discusses the fault-trajectory approach suitability for fault diagnosis on analog networks. Recent works have shown promising results concerning a method based on this concept for ATPG for diagnosing faults on analog networks. Such method relies on evolutionary techniques, where a generic algorithm (GA) is coded to generate a set of optimum frequencies capable to disclose faults.*

## 1. Introduction

The modern fabrication processes for electronic circuitry allow highly integrated devices. Such possibilities together with the endless wishes of customers drive the electronic industry, demanding good solutions for test, design for testability and diagnosis. Such wishes boosters the complexity of the test issue. The diagnosis problem may be considered the paramount, and, so far, there is a lack in the literature of an approach generic enough, to be considered a method. Our issue intends to discuss the suitability of a *fault trajectory* approach for component fault diagnosis on analog circuits. The method is based on *the fault trajectory concept* for fault diagnosis on analog networks, which relies on evolutionary techniques, where a GA is coded to optimize test vector generation.

## 2. The method

The method consists of finding a convenient test signal to excite the circuit under test (CUT), and observe at output, a response, that may be considered faulty, depending on a given requirements. According to the behavior of the output response faulty components diagnose can be performed. The test signal must have the following requirements: it must disclose faults in the circuit, and, in addition, it must disclose which component is faulty also. However, on the other hand, it shouldn't have an excessive number of different input patterns. Then, the test may not consume a great amount of time; nor should generate a test vector long and complex, which may be unfeasible and expensive. Yet, at the circuit's output, the response signal shall be convenient for a detection issue. For such needs, GAs [1], as robust search-space data miners, seems to be a very useful approach to discover potential set of optimal solutions [2,3], GAs can be used to produce test patterns [4]. In this case, GAs are able, from a first set of random test patterns generated automatically, to verify the suitability of each pattern towards the ideal solution defined by the user [5]. Some of these patterns may be discarded and others may be recombined among each other, in a continuous search for even better test patterns.

### 2.1 The fault model and the fault dictionary

Our approach uses a fault simulation process (FS), where an adequate fault model must be used. It is assumed that the faults are under *the functional **parametric fault*** [7] paradigm: a fault in a circuit will be the result of a parametric deviation in a component value. This way, faults in R & C are represented as % deviations on their values, and faults on active devices will be represented as % deviation on the values of their macro model [7]. The FS process consists of building from the original circuit (the one with nominal values components) a set of different circuits (the faulty ones), where their components are out the nominal value at a time. This is done inserting faults on all its components (systematic % deviation on its values) within a given range. As an example if the target circuit is a filter, and its main requirement is its frequency response, one can assume that a good attempt for a test vector is a set of sinusoidal signals. Of course, the filtering behavior is totally checked against a frequency sweep generator, but, in practice, such approach is unfeasible especially for built-in structures. This way, what we look for is a test vector composed by a minimum of frequencies. The correct choice of such set of signals is made via the use of a FS and a GA for optimizing the search, since the number of faulty circuits against a possible set of frequencies are infinite.

### 2.2 Application example

We have used a normalized biquad negative feedback low-pass filter as the CUT[7]. The devices of the CUT whose faults are to be detected are the seven passive components. When a parametric fault is inserted, the value of a passive device is transformed in a value, which may range from e.g. 60%-140% of its nominal value. For each passive component faulty circuits are generated using parametric deviations in steps of 10% of deviation, the zero % represents the nominal behavior (fig.1). High diagnosis will be possible if our test vector is able to distinguish a faulty device from another one. This way, we have to look for the test vector components that are able to evidence every faulty component influence in the circuit response. Using the frequency-sampling theorem, we propose a linear transformation that reduces all observable spectra of a given response to a point of a Cartesian coordinate space. Figure 2 depicts 2 curves, the 1[st] one is called H, and represents the golden *behavior*, and the 2nd one, called K, a *faulty* response due to a defective component. If both circuits were stimulated by a test vector composed by, e.g. two frequencies $f_1$ & $f_2$, it will be equivalent to sample the two curves at $f_1$ and $f_2$ frequencies. The sampling performed in fig.2 leads to the following transformation into a Cartesian coordinate XY plane: $H(f_1)=A_1$, $K(f_1)=B_1$, and $H(f_2)=A_2$ $K(f_2)=B_2$. As the sampling frequencies are the same for each curve, the transformation yields the pairs: *(A1,A2)*, & *(B1,B2)*, as *XY* coordinate points. The 1st point represents the golden behavior and the 2nd a faulty one. Some simplification is introduced if we consider the *golden behavior* point as the Cartesian coordinate plan origin.

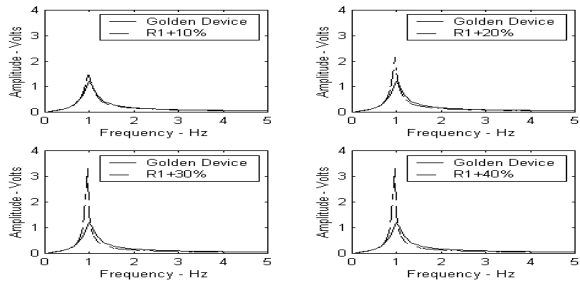

**Figure 1: "golden behavior" & "fault dictionary items".**

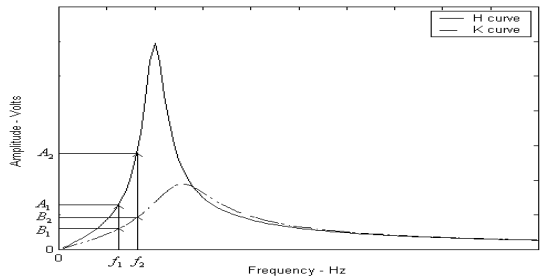

**Figure 2: the transformation into a coordinate data.**

### 2.3 The "Fault Trajectory"

As we are dealing with analog linear continuous-time circuits, we consider reasonable the responses are smooth and monotonic. Crescent/De-crescent parametric deviations on components within a given range shall produce a set of coordinate points in the plane which can be connected, to compose what we define a fault trajectory. According to this concept, each deviated device in the circuit has its own trajectory, or "component parametric fault trajectory", in that Cartesian plan. Fig.3 depicts such situation, in this case for the component R3[7]. Each fault trajectory, in order to be distinguished each other, must have a totally independent "pathway". Finally, the "efficacy", or diagnosis grade, for a given test vector, is enunciated as: "One shall be able to find out a set of frequencies $f_1, f_2,...f_n$ to compose a test vector in order to distinguish the highest *number of fault components via circuit faulty behavior*". To accomplish the above statement (i.e., to find out the frequencies that composes the test vector), evolvable techniques are used. This optimizes the search of the best solution over a spread and wide universe of possible solutions.

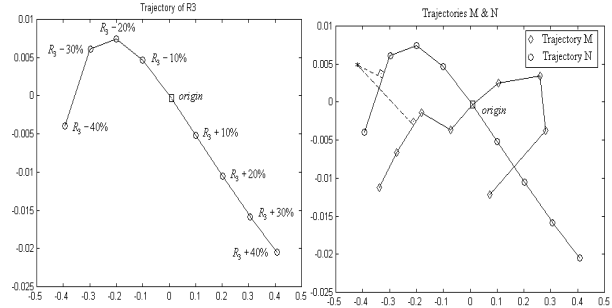

**Figure 3: R3 fault trajectory (left), fault diag. (right).**

### 2.4 The "Genetic Algorithm"

We have chosen a test vector composed by two frequencies. A GA performs the suitability analysis of a given test vector. The basic flow of the GA is found in [6]. Its main features are: *128* individuals, *15* generations, *reproduction rate* of *50%*, *mutation rate* of *40%*, the "*roulette wheel*" as the mining method, and the *number of generations* as the stop criteria. The selection method or the *fitness* criteria are based on the search of a graphical configuration for the trajectories that minimizes the number of common pathways, and intersections among the fault trajectories. The fitness of a test vector composed by $f_m$, $f_n$ is given as: ***fitness ($f_n$, $f_m$) = 1 / (1+I)***. Where *I* is the number of trajectories intersections. Each faulty circuit corresponds to a point in our *XY* plane. These points can be piece-wised, and considering that just one circuit's component is faulty at a time, we have the procedure: Given a point in the Cartesian plane due to an unknown fault, it can be assigned to a PW segment, which would be the segment with the highest probability to be the right one. Such operation is done drawing perpendiculars from known fault trajectories to the point where the unknown fault is (fig 3) the unknown fault is represented by an (*). In this example, the unknown fault seems to belong to the N device trajectory, since its distance to the only two segments from which perpendiculars exist is smaller for the N-type than to the M-type fault.